%% file: ms.tex
\documentclass[runningheads]{llncs}
\usepackage{graphicx}
\usepackage{color}
\usepackage{wrapfig}
\usepackage[caption=false]{subfig}
\usepackage[hidelinks]{hyperref}
\usepackage[T1]{fontenc}
\usepackage[misc,geometry]{ifsym}

\begin{document}

\newcommand\missing{{\color{red}\textbf{[??]}}}
\makeatletter
\newcommand\footnoteref[1]{\protected@xdef\@thefnmark{\ref{#1}}\@footnotemark}
\makeatother

\title{Hospitalization Length of Stay Prediction using Patient Event Sequences}

%
%
\author{Emil Riis Hansen\inst{1}(\Letter)\orcidID{0000-0003-4103-1244} \and
Thomas Dyhre Nielsen\inst{1}\orcidID{0000-0002-4823-6341} \and
Thomas Mulvad\inst{2}\orcidID{0000-0002-4457-3965} \and
Mads Nibe Strausholm\inst{2}\orcidID{0000-0002-4099-3209} \and
Tomer Sagi\inst{1}\orcidID{0000-0002-8916-0128} \and
Katja Hose\inst{1,3}\orcidID{0000-0001-7025-8099}}
\authorrunning{Hansen et al.}
%
\institute{Department of Computer Science, Aalborg University, Aalborg, Denmark
\email{\{emilrh,tdn,tsagi,khose\}@cs.aau.dk}\\
\and
Unit of Business Intelligence, North Denmark Region, Aalborg, Denmark
\email{\{tml,mns\}@rn.dk}
\and
TU Wien, Vienna, Austria\\
\email{katja.hose@tuwien.ac.at}
}

\maketitle              
\begin{abstract}
Predicting patients' hospital length of stay (LOS) is essential for improving resource allocation and supporting decision-making in healthcare organizations. This paper proposes a novel approach for predicting LOS by modeling patient information as sequences of events. Specifically, we present a transformer-based model, termed Medic-BERT (M-BERT), for LOS prediction using the unique features describing patients' medical event sequences.  
We performed empirical experiments on a cohort of more than $45k$ emergency care patients from a large Danish hospital. Experimental results show that M-BERT can achieve high accuracy on a variety of LOS problems and outperforms traditional non-sequence-based machine learning approaches.  

\end{abstract}

\input{sections/introduction.tex}
\input{sections/related.tex}
\input{sections/methodology.tex}
\input{sections/data.tex}

\input{sections/results.tex}

\input{sections/conclusion.tex}

\section*{Acknowledgments}
This work is partially supported by the Poul Due Jensen Foundation and the Region North Denmark Health Innovation Foundation.

\bibliographystyle{splncs04}
\bibliography{bibliography}

\end{document}

%% file: sections/introduction.tex
\section{Introduction}

Increasingly scarce hospital resources challenge (often oversaturated) hospital wards, with a negative impact on the quality of health care at the hospitals~\cite{af2020association}
Models for predicting the remaining time of patient admissions, i.e., patient length of stay (LOS), could be invaluable for healthcare facilities to plan the availability of beds, staff, and other essential resources. For instance, automatic prediction of discharge time could be used in administrative planning systems for preemptively freeing in-hospital resources to alleviate hospital ward oversaturation~\cite{stone2022systematic}. However, LOS prediction is a challenging problem, requiring methods for handling missing data~\cite{sha2017interpretable} and integration of temporal event dependencies.

Previous work on LOS prediction models patient hospitalizations using tabular data with imputation techniques for replacing missing values~\cite{bacchi2022machine}. 
While tabular data is the most common data representation in Machine Learning (ML) models, it has several drawbacks. Among others, it does not provide immediate support for integrating the temporal dependencies between observations, such as the order of treatments, or the time of conducted procedures. Moreover, standard ML techniques for tabular data, such as artificial neural networks (ANN), gradient boosting (GB), and support vector machines (SVMs), require complete data, hence often relying on imputation techniques when data is incomplete. However, missing data observations in healthcare data are often not missing at random (NMAR), meaning that the mere fact that an observation is missing is in itself important information~\cite{li2021imputation}.

To alleviate the problem of temporal dependencies and missing data, attention models, also known as transformers, have recently been investigated for Electronic Health Record (EHR) data formatted as sequences of medical events~\cite{lequertier2021hospital}. However, embedding-based transformer approaches have, to the best of our knowledge, not previously been applied to LOS prediction. This work examines how attention models, can be utilized for this task.

Attention models using self-attention alleviate the inefficiency of recurrence networks for long sequences~\cite{song2018attend}. However, they still capture significant sequential information by learning from the order of tokens in the sequence. In medical data, multiple observations may be given the same timestamp, with no meaning assigned to their individual order within the corresponding event. For example, a blood panel drawn from a patient contains several individual measurements whose internal order is insignificant. 
Based on layers of transformer encoders, we propose a revised attention model, henceforth called Medic-BERT (M-BERT), based on the Natural Language Processing (NLP) model BERT~\cite{bert19} and its revised semi-supervised training method. We employ the model for LOS prediction based on sequences of patient specific medical events happening during hospitalization and which exhibit the event concurrences common in patient data. We evaluate our method on a cohort of more than $45k$ patient admissions from a large Danish hospital with diverse medical events, such as vital measurements, medication administration, laboratory tests, and conducted procedures.

The rest of this paper is structured as follows. Section \ref{sec:related} reviews related work on LOS prediction. Section \ref{sec:method} presents our model for representing hospitalizations as event sequences and our proposed M-BERT model for this unique data. Section \ref{sec:data} describes the large dataset used to evaluate this work and Section \ref{sec:eval} the evaluation and its results. We conclude with Section \ref{sec:conclusion}. This work is an extended version of Hansen et al.~\cite{aime2023hansen} published at AIME 2023. 



%% file: sections/related.tex
\vspace{-1em}
\section{Related Work}
\label{sec:related}

The LOS prediction problem can be stated in different ways varying in the required resolution of the prediction, applied methods, and structure of patient data.
Iwase et al.~\cite{iwase2022prediction} investigate the use of Machine Learning (ML) methods for LOS prediction of intensive care unit patients (ICU). Using Random Forest (RF), Gradient Boosting (GB), and Artificial Neural Networks (ANN) technologies, on binary stratification of patient admissions into long (more than one week) and short (less than one week) stays. RF showed high predictive performance. In contrast to the tabular structure of their data, we instead attempt to utilize the temporal dependencies within the data by structuring the patient data as medical event sequences. In Batista et al.~\cite{batista2020methodology}, patients are stratified into three categories (LOS $< 3$, $3 <$ LOS $< 10$, LOS $>= 10$) using RF and Support Vector Machines (SVM). The highest performance was achieved for RF models. While the nature of their ML models requires manual feature selection and missing data imputation, our method does not require any feature selection for in-hospital measurements, nor does it rely on data imputation. 

The most challenging LOS prediction task is regression-based LOS, where the precise hospitalization LOS as a real value is to be predicted. Barsasella et al.~\cite{barsasella2022machine} investigated a range of classical AI models, such as Decision Trees (DTs), RFs, Logistic Regression (LR), and SVMs for real-valued LOS prediction. While standard ML models work for tabular structured patient data, they neglect the important temporal dependencies between in-hospital patient events. While much work has been done in LOS prediction and regression using standard ML methods, only a few works attempt the usage of sequence models.

Seminal work in the area of transformer models for patient sequence data includes that of Li et al.~\cite{li2020behrt} and Rasmy et al.~\cite{rasmy2021med}, where sequences of diagnosis codes for consecutive hospitalizations are used as input to a BERT like transformer model for diagnosis code prediction tasks. While Rasmy et al.~\cite{rasmy2021med} investigate prolonged LOS $> 7$ days event prediction as a model pre-training task, they do not predict future LOS for new patients. Furthermore, while their approaches are only investigated for sequences consisting of diagnosis codes, we evaluate sequences consisting of multiple patient event types. 

In Meng el al.~\cite{meng2021bidirectional}, a transformer model is used to predict the onset of depression. They integrate various medical types for patient event sequences, including diagnosis and medication events. While their sequences include code-based medical events, we extend with medical measurement events, such as laboratory tests and vital measures while encoding the event value as part of the event token.

The work closest to ours is that of Song et al.~\cite{song2018attend}, where an attention model is used to predict categorical LOS based on patient sequences. Whereas their approach targets dense time-series data, where the same measurements are present at every timestep, instead, we investigate a learned embedding approach where the input sequences consist of codes with different types of medical events.       




%% file: sections/methodology.tex
\section{Transformer Models for EHR}
\label{sec:method}
In this section, we describe how EHR data can be modeled as medical event sequences and our usage of transformer models for learning from these sequences. 

\subsection{Hospitalizations as Event Sequences}\label{subsec:event_sequences}
\begin{figure}[tb]
    \centering
    \includegraphics[trim={1cm 11.5cm 8cm 0cm}, width=0.9\linewidth]{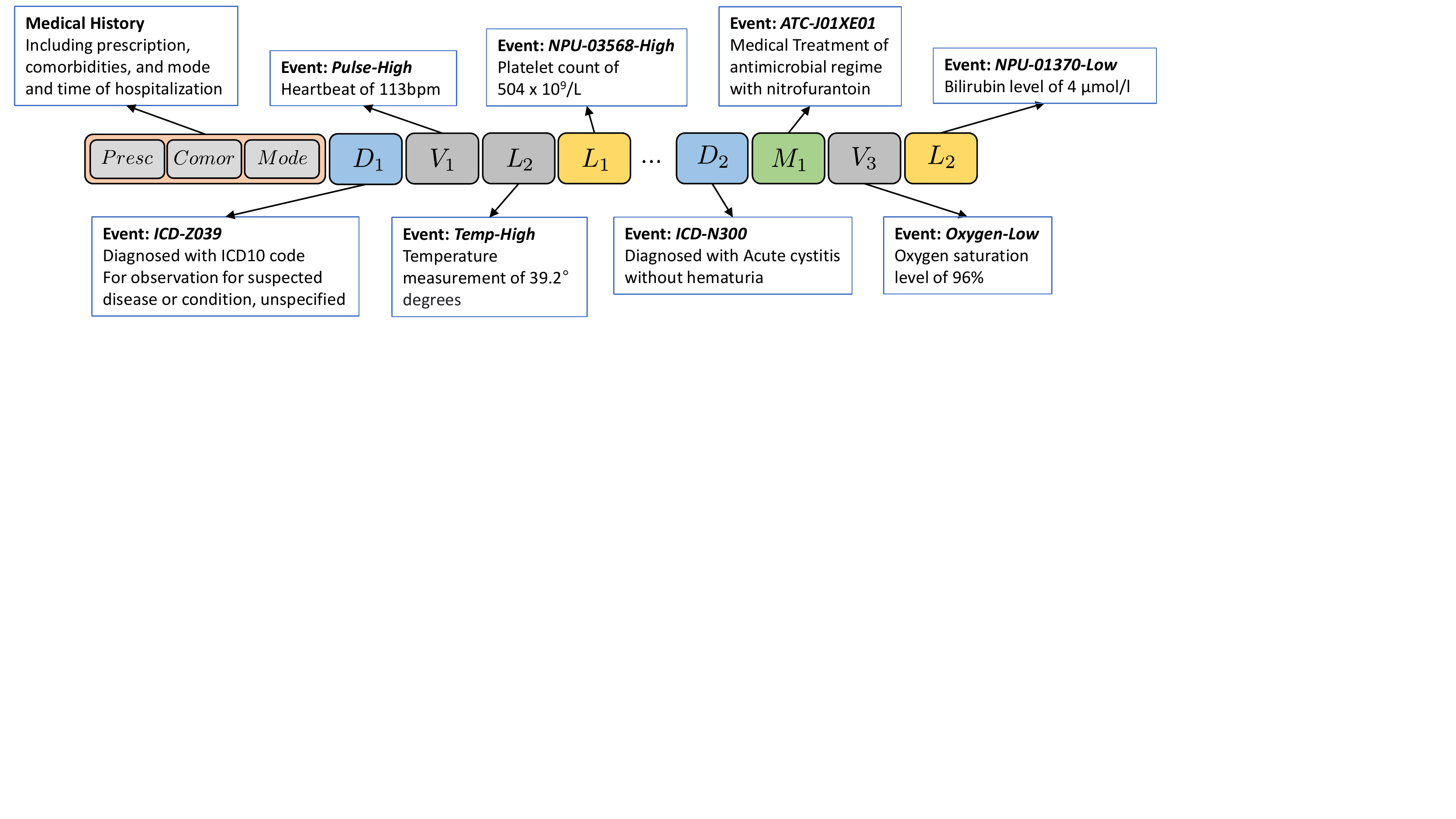}
    \caption{Illustration of a patient event sequence. Starting with the patient's medical history as summarized in Table \ref{tab:medical_history}, the patient is initially diagnosed with the ICD-10 code \emph{Z039}. Subsequent vital measurements and laboratory tests are performed to monitor the patient's state and determine the patient's underlying condition. Consequently, the patient is diagnosed with acute cystitis without hematuria (ICD-10 code \emph{N300}), and antibiotic treatment is initiated with nitrofurantoin (ATC code \emph{J01XE01}). After additional procedure and therapeutic medical events, the patient is released from the hospital.} 
    \label{fig:patient_sequence}
\end{figure}

Patient hospitalizations can be naturally modeled as sequences of medical procedures for determining, measuring, or diagnosing the patient's condition. Other medical procedures are therapeutic and intend to treat or cure the patient. To standardize how medical procedures are described, medical facilities code procedure concepts using accepted medical taxonomies often used in ML applications~\cite{HansenSHLLS22}, such as the Anatomical Therapeutic Classification (ATC)~\cite{ronning2002historical} for medication administration and the International Classification of Diseases and Related Health Problems (ICD)~\cite{cartwright2013icd} for condition diagnosis. Hence, a patient hospitalization can be described as the sequence of concept tokens detailing medical procedures pertaining to a patient coded using medical concepts from accepted medical taxonomies. An example patient hospitalization sequence is illustrated in Figure \ref{fig:patient_sequence}.       

Furthermore, as a patient's medical history is crucial for correct management and treatment, we pre-pend the patient's medical history to the hospitalization sequence as a tokenized vector as illustrated in Figure~\ref{fig:patient_sequence}. The vector consists of 38 tokens describing the patient's medical history, including comorbidities from the Charlson Index~\cite{sundararajan2004new}, five years of medications prescription history grouped by the first level of the ATC hierarchy~\cite{ronning2002historical}, and the mode, time, and initial triage category~\cite{wireklint2021updated} of hospitalization. The historical information included in the vector is summarized in Table~\ref{tab:medical_history}.    

\begin{wraptable}{r}{5.5cm}
\vspace{-2em}
    \centering
    \caption{Patient medical history.}
    \label{tab:medical_history}
	\begin{tabular}{p{3.5cm}p{1.7cm}}
        \textbf{Data}  & \textbf{\#Tokens} \\
        \hline
        Comorbidities & 18   \\
        Prescription history & 14  \\
        Mode, time, \& triage &  6    
	\end{tabular}
    \vspace{-2em}
\end{wraptable}

\subsection{Hospital Measurement Events}\label{subsec:event_measurements}

For some medical procedures, such as vital measurements and laboratory tests, a numerical measurement value accompanies the procedure. While other works in transformer models for EHR data disregard the numerical values of measurements~\cite{meng2021bidirectional,rasmy2021med,li2020behrt}, we instead integrate this information as part of the patient input sequences because numerical measurement values add important information regarding the state of a patient. For example, the knowledge that a temperature measurement was performed is naturally important information. Still, from the measurement value of $40.1^{\circ}C$, we learn that the patient has a fever. Using patient-specific threshold values for measurements based on age, gender, and pregnancy status, we map measurement values into tokens representing either normal, abnormal-low, or abnormal-high findings. For example, given that we measure an albumin level of $56$ g/L for a $31$-year-old male patient, we would create the token \emph{albumin-high} to reflect that the value of the measurement was above what is expected ($36$-$48$ g/L) for a patient with the given demography.  


\begin{figure}[bt]
    \centering
    \includegraphics[trim={1.7cm 12cm 11.7cm 0cm}, width=0.9\linewidth]{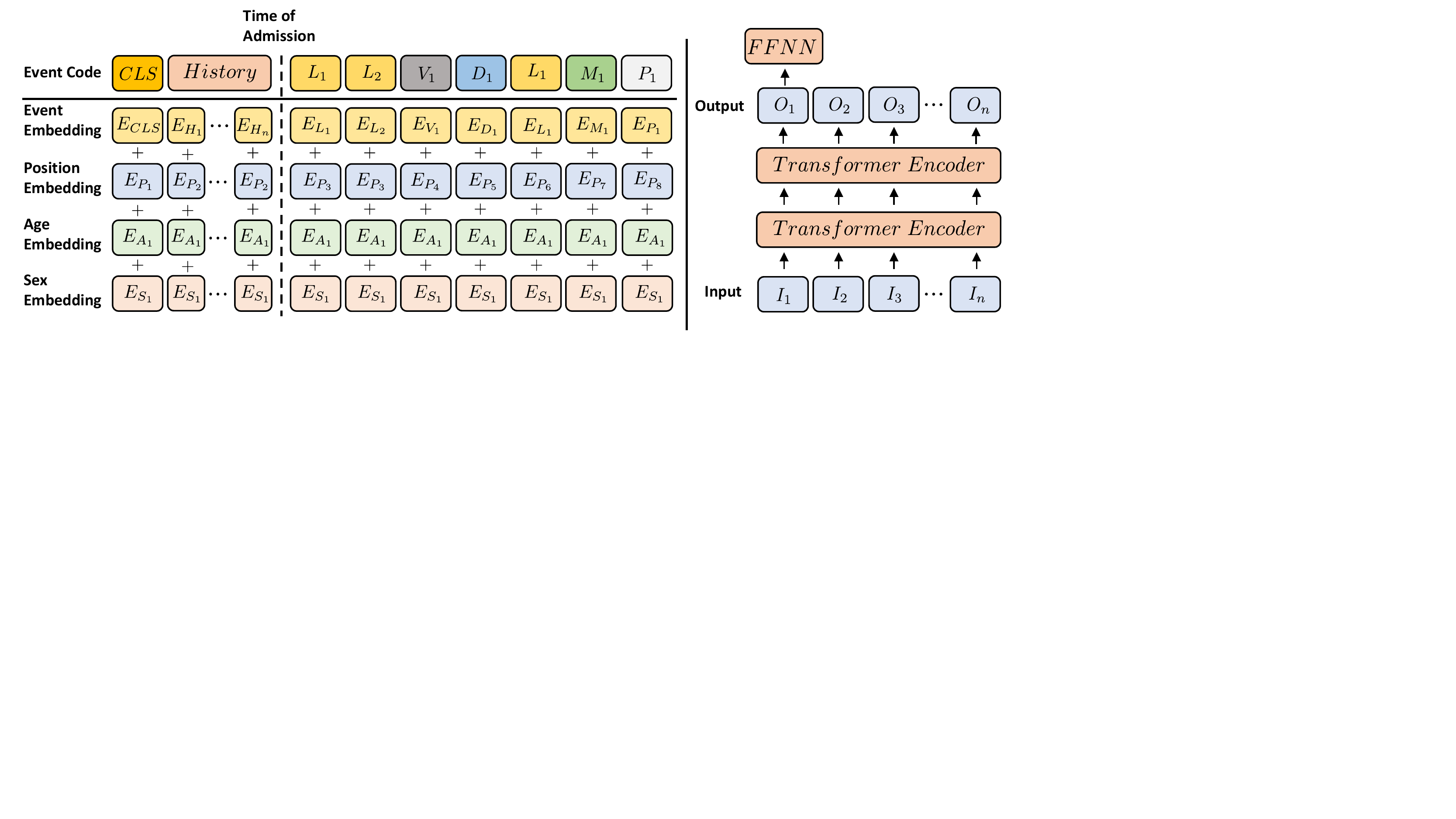}
    \caption{Medical event sequence pre-pended with the patient's medical history.} 
    \label{fig:sequence_model}
\end{figure}

\vspace{-1em}
\subsection{Transformer Models for EHR Data}
To investigate the challenge of LOS, we examine a modified version, henceforth termed Medic-BERT (M-BERT), of the Bidirectional Encoder Representations from Transformers (BERT)~\cite{bert19} model for EHR data. BERT is an NLP model based on a stack of encoder layers from the transformer architecture introduced by Vaswani et al.~\cite{vaswani2017attention}. We argue that sequence models, such as BERT exhibit properties beneficial for solving medical tasks based on EHR data. The transformer encoder naturally handles the complex long-term dependencies that occur between medical concepts through its utilization of multi-head self-attention. BERT can naturally integrate disparate modalities, such as diagnostic end therapeutic events, as each event is encoded as an n-dimensional vector token. Furthermore, BERT naturally operates in domains with irregular intervals between events, as is the case with EHR data. We, therefore, investigate our modified version of BERT for LOS prediction for patient event sequences. The M-BERT architecture is illustrated in Figure \ref{fig:sequence_model}.

M-BERT learns an embedding for each medical event token while trained toward LOS prediction. The position embedding enables the model to learn from the temporal dependencies within a sequence. We use a static position embedding as described in Waswani et al.~\cite{vaswani2017attention}; however, modified for the usage on medical event sequences as described in Section~\ref{subsec:position_embedding}. As patient demographics are a vital part of any medical prediction model, we pay special attention to this information by adding a trainable age and sex embedding at each medical event~\cite{li2020behrt}. Furthermore, as in the original BERT model, we use the special classification (CLS) token as a final aggregate representation of the sequence. Hence, as illustrated in Figure~\ref{fig:sequence_model}, the CLS representation is fed to a linear output layer for LOS classification and regression tasks.          

\subsection{Position Embedding}\label{subsec:position_embedding}
\begin{figure}[tb]
    \centering
    \includegraphics[trim={1cm 14cm 14cm 0cm}, width=0.9\linewidth]{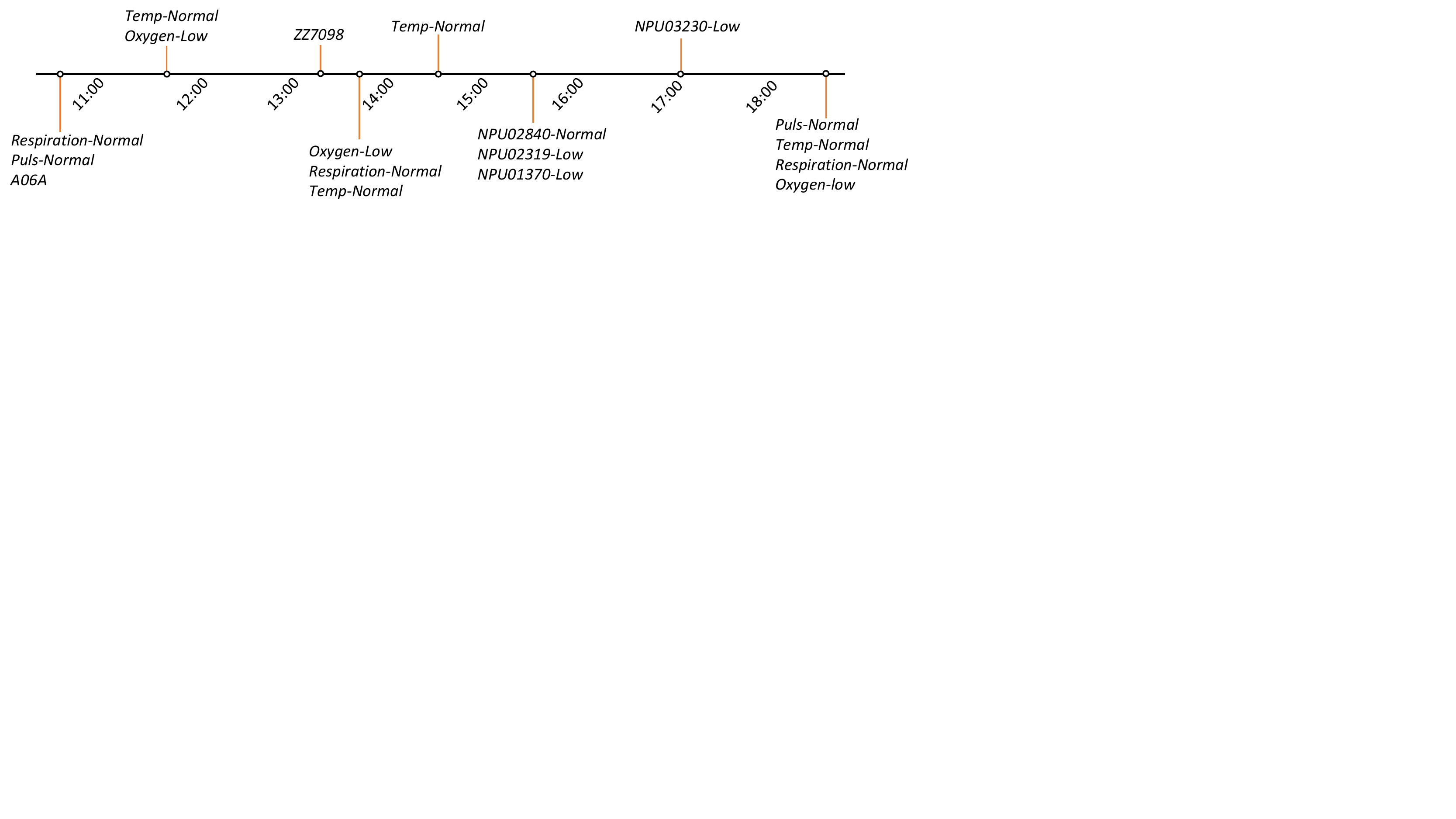}
    \caption{Patient event sequence illustrating events grouping together.} 
    \label{fig:event_chunks}
\end{figure}

Due to the nature of patient care and hospital administration, some measurement events tend to chunk together. Clinicians will often conduct various measurements, such as blood pressure, temperature, and heart rate, over a short time and later persist the information into the patient's EHR. Hence, we are sometimes prevented from knowing the specific times and order of medical events. This effect is most frequent for vital measurements and laboratory tests. Multiple laboratory tests are often conducted on the same patient sample, such as a single blood sample, making it impossible to know the chronological order for such measurement events. An example sequence of patient events chunking together is illustrated in Figure \ref{fig:event_chunks}. To enable the model to understand that these events have no fixed ordering, we assign the same position embedding to events co-located in time as illustrated in the position embedding of Figure \ref{fig:sequence_model}. For example, the $L_1$ and $L_2$ events are both mapped to the $E_{P_3}$ position embedding.

%% file: sections/data.tex
\section{Data}
\label{sec:data}
\begin{figure}[tb]
  \centering
  \subfloat[Histogram of length of stay.]{\includegraphics[width=0.47\textwidth]{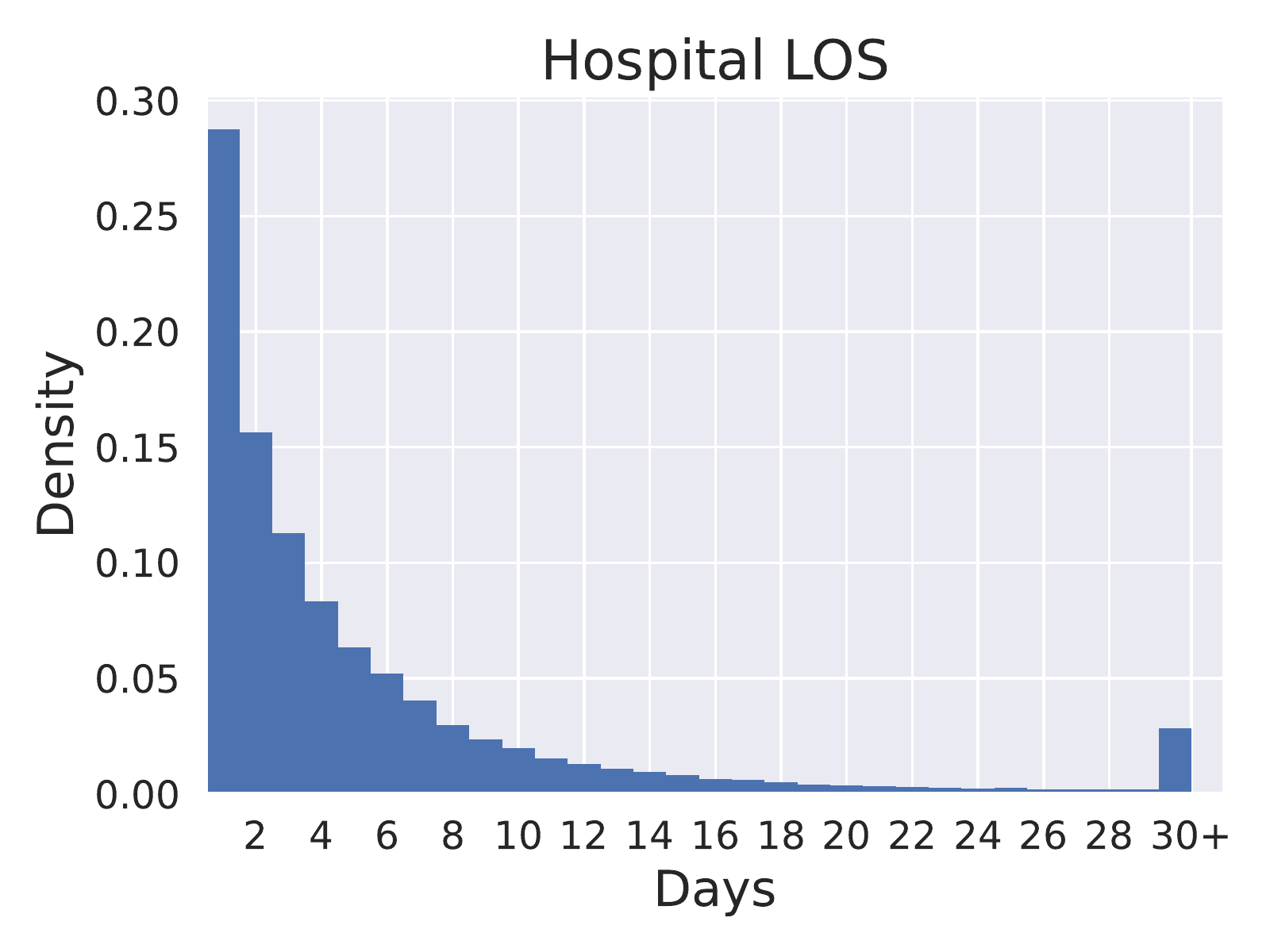}\label{fig:los_hist}}
  \hfill
  \subfloat[Categorical length of stay.]{\includegraphics[width=0.47\textwidth]{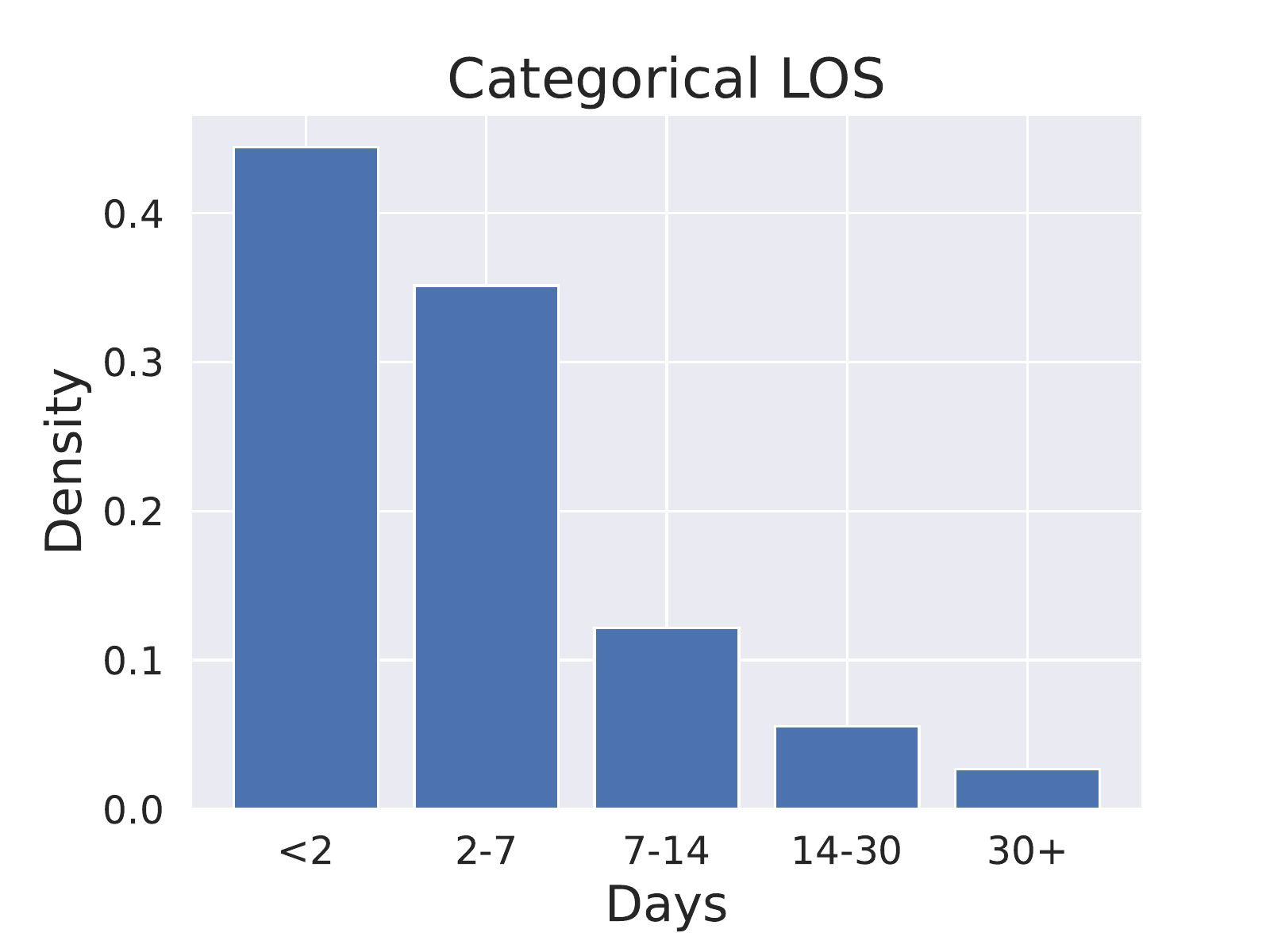}\label{fig:los_cat}}
  \caption{Illustrations for length of stay distribution over patient hospitalizations.}
  \label{fig:los_histograms}
\end{figure}

We compiled a Danish dataset of patients admitted to a large hospital in northern Jutland from the period 2018-2021 to investigate transformer models for LOS prediction. The dataset consists of $48,177$ emergency care patients with admissions longer than one day. Figure~\ref{fig:los_histograms} illustrates the distribution of the remaining length of stay for the patient cohort. In a clinical setting of resource allocation for emergency care patients, we are mostly interested in the planning of patient care for a couple of weeks in advance. Hence, we clip long patient stays to 30 days of admission to better fit and optimize for the clinical setting, as we are not interested in precisely predicting the long-tail distribution of the data. 

In Denmark, each person can be uniquely identified by an identification number from the central person register (CPR) henceforth referred to as a CPR-number. As medical events are each associated with a unique CPR-number, it is possible to connect the medical data pertaining to a patient from disparate databases. As mentioned in Section~\ref{subsec:event_sequences}, we divide patient medical data into historic information and admission-specific information. Historic information includes prescriptions, comorbidities, and mode and time of hospitalization, whereas admission information includes laboratory tests, vital measurements, hospital-administered medicine, and procedure codes.

\begin{wraptable}{r}{6.6cm}
    \caption{Concept types with their occurrences in the dataset.}
    \label{tab:concept_types}
	\begin{tabular}{p{2.5cm}p{1.7cm}l}
        \textbf{Event Type}  & \textbf{Tokens} & \textbf{Occurrences}         \\ \hline
            Lab Tests      & $748$                         & $2,774,790$ \\
            Vital Measures & $22$                          & $837,931$ \\
            Medication     & $1,441$                       & $376,591$ \\
            Procedures     & $2049$                        & $247,924$ \\
            History        & $81$                          & $1,880,580$	
        \end{tabular}
\end{wraptable}

Danish laboratory tests are coded using the Nomenclature for Properties and Units (NPU) terminology~\cite{npu2009}. NPU ensures that laboratory tests are standardized and patient examinations can be used and understood by all clinicians. 
As detailed in Section~\ref{subsec:event_measurements}, we integrate the semantics of laboratory test results as part of the event token by appending either high, low, or normal with respect to the patient demographics to the laboratory event token, thus encoding the meaning of the result. Vital measurements are extracted from a system called Clinical Suite and consist of the 7 most common vital observations, including temperature, oxygen saturation, BMI, pulse, respiration rate, blood pressure (systolic and diastolic), and oxygen supplement. As with laboratory tests, patient-specific thresholds are used to encode the meaning of the result into event tokens. In-hospital administered medication is coded using the ATC taxonomy~\cite{ronning2002historical} and consists of more than $5$k chemical substances. Procedure codes specify in-hospital procedures performed on patients. While diagnosis events are an important medical modality, the dataset does not contain the time of such events. Hence, we omit diagnosis events from patient sequences. The event types with distinct tokens and total occurrence in the data are summarized in Table~\ref{tab:concept_types}.

%% file: sections/results.tex
\section{Empirical Evaluation and Results}\label{sec:eval}
In this section, we explain the experimental settings and results of the empirical evaluation.

\subsection{Experimental Setting}
Transformer models are trained, as a rule, using unsupervised pre-training for learning general token embeddings, followed by supervised fine-tuning targeting a specific downstream task. As we are only focusing on a single prediction task, we directly train model parameters and token embeddings toward the downstream task of LOS prediction without pre-training. The experimental code is available online\footnote{\label{footnote:online_appendix}\url{https://github.com/dkw-aau/medic\_transformer}}.

We use the admission data gathered within the first 24 hours of admissions for LOS predictions. Hence, we remove sequence events happening after 24 hours of admission. We evaluate our approach on three LOS experiments of increasing complexity. The first two experiments are a \textbf{Binary} classification of LOS $> 2$ days, and a three-class \textbf{Category} task 
of LOS $> 2$, $2 \leq$ LOS $\leq 7$, and LOS $> 7$ days 
with class balances as illustrated in Figure \ref{fig:los_cat}. The last experiment, termed \textbf{Real}, is a regression task with the objective of predicting the LOS as the real number, with a histogram of admission times as illustrated in Figure~\ref{fig:los_hist}.

We evaluate our approach against three ML models: RF, ANN, and SVM. We use implementations of the models from the Sklearn library~\cite{scikit-learn} with standard model hyperparameters. In preparing samples for these models, we use the latest measured value for each event type (within 24 hours of admission) as input features~\cite{iwase2022prediction}. Subsequently, the mean of variables is used for imputation of missing data values, and variables are scaled to values between 0 and 1. Lastly, a chi$^2$ test is used in feature selection for the selection of the 50 most relevant features.     

Our model is trained on a random split distributed as 80/10/10 of all patient samples for training, validation, and testing. We use the loss of the evaluation data for early stopping training if the loss does not go down within ten training epochs. The model architecture has six hidden layers with an intermediate layer size of 288, eight attention heads, and input token embeddings have a size of 288. We truncate sequences to 256 tokens, as most sequences adhere to this limit. To counter overfitting, we add a dropout layer with a probability of 10\% after the output of the final encoder layer, attention dropout at every layer with a probability of 10\%, and weight decay of $0,003$. Furthermore, experiments were performed with a learning rate of 1e-5.
 


\begin{figure}[tb]
  \centering
  \subfloat[AUROC curves for the \textbf{Binary} LOS experiment over all methods.]{\includegraphics[width=0.47\textwidth]{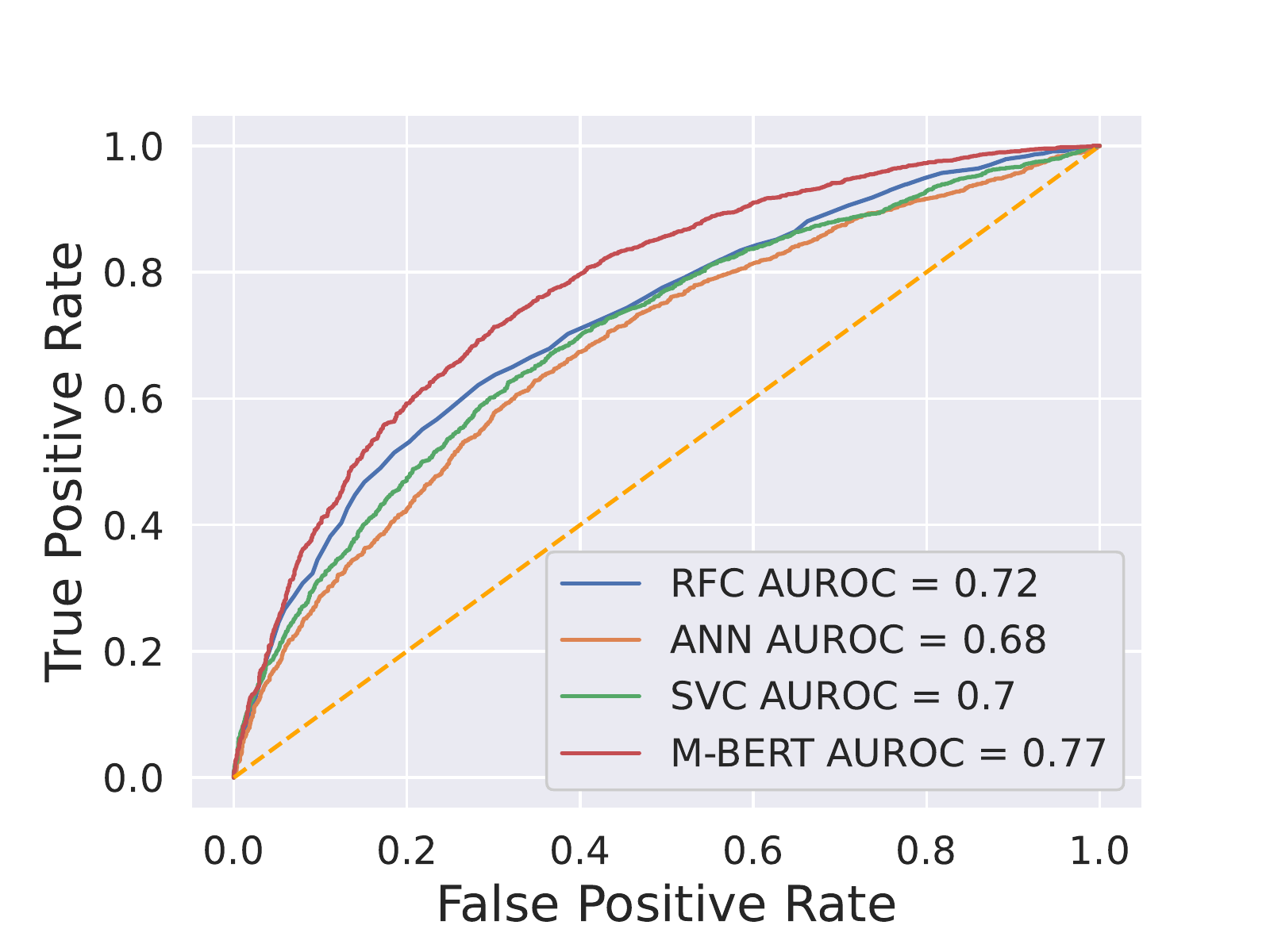}\label{fig:roc_all}}
  \hfill
  \subfloat[\textbf{Binary} experiment AUROC curves for different age groups over M-BERT.]{\includegraphics[width=0.47\textwidth]{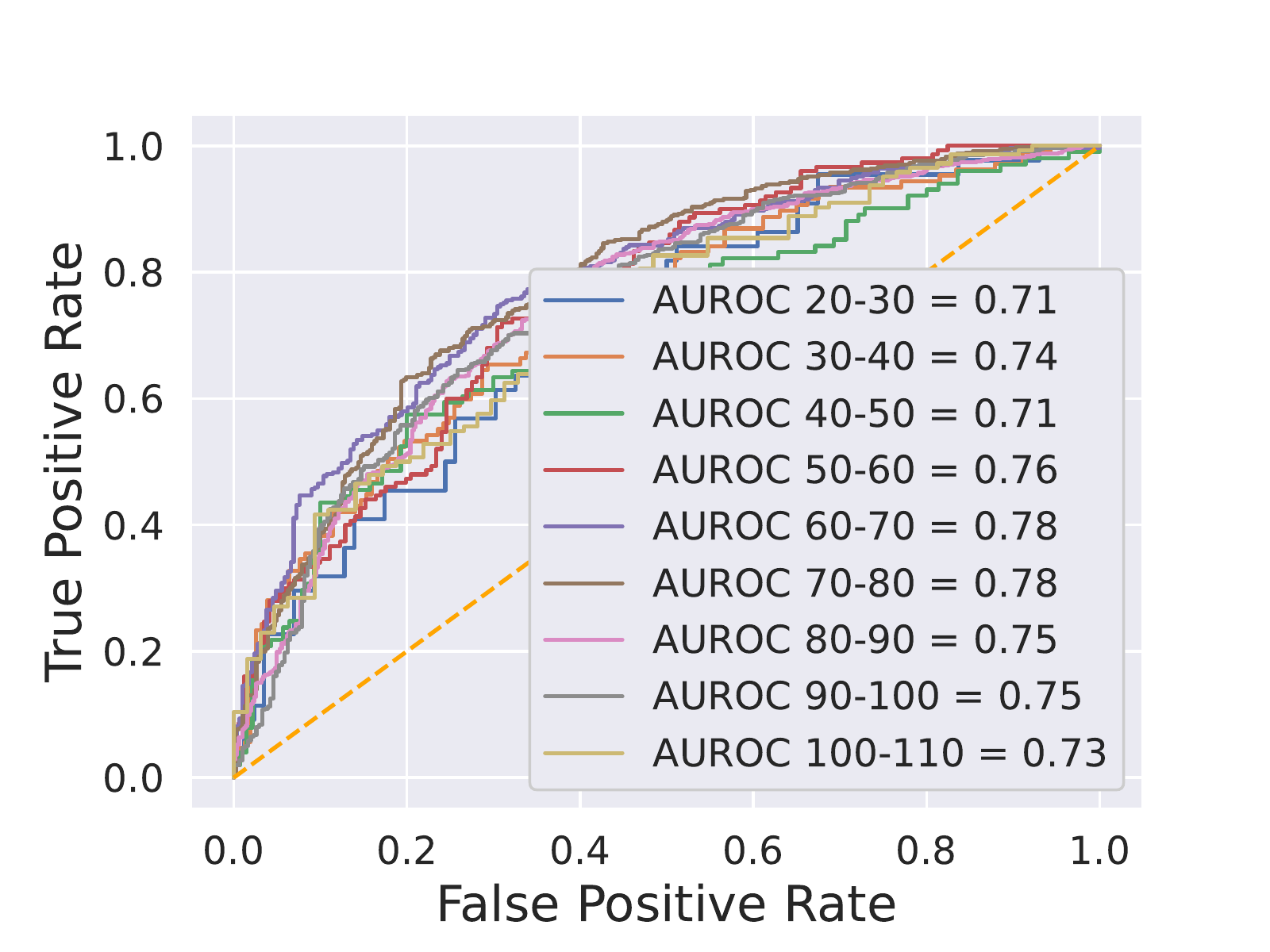}\label{fig:roc_ages}}
  \caption{AUROC plots for experimental results on the binary prediction task.}
  \label{fig:auroc_plots}
\end{figure}

\vspace{-1em}
\subsection{Results}

Table \ref{tab:results} presents the Area Under the Receiver Operating Characteristics \mbox{(AUROC)} and harmonic mean of precision and recall (F1) values for the \textbf{Binary} and \textbf{Category} experimental settings and Mean Absolute Error (MAE) and Mean Squared Error (MSE) for the LOS regression task. Furthermore, Figure~\ref{fig:roc_all} presents the AUROC curves for the \textbf{Binary} experimental setting. M-BERT outperforms the traditional ML techniques in all experimental settings. The results indicate that transformer models may be able to leverage the temporal dependencies inherent to patient EHR data for increased predictive accuracy. 
\begin{wraptable}{r}{9.2cm}
\caption{Experimental results.}
    \label{tab:results}
    \begin{tabular}{lccccccccc}
    \hline
    \textbf{} & \multicolumn{2}{c}{\textbf{Binary}}  & \textbf{} & \multicolumn{2}{c}{\textbf{Category}} & \textbf{} & \multicolumn{2}{c}{\textbf{Real}}     \\ 
    \cline{2-3} \cline{5-6} \cline{8-9}
    \textbf{} & \textbf{AUROC} & \textbf{F1} &           & \textbf{AUROC} & \textbf{F1}  &           & \textbf{MAE} & \textbf{MSE} \\ 
    \hline
    RF        & $0.72$ & $0.70$ &    & $0.66$ & $0.45$ &    & $4.18$ & $39.08$ & \\ 
    ANN       & $0.67$ & $0.68$ &    & $0.63$ & $0.43$ &    & $4.09$ & $38.10$ & \\
    SVM       & $0.70$ & $0.70$ &    & $0.65$ & $0.38$ &    & $3.56$ & $43.36$ & \\
    \hline
    M-BERT  & \textbf{0.78} & \textbf{0.77} &    & \textbf{0.74} & \textbf{0.54} &    & \textbf{3.42} & \textbf{37.48} & \\ 
    \hline
    \end{tabular}
\end{wraptable}
Furthermore, being a transformer-based model, M-BERT overcomes the challenge of missing data and imputation, as patient sequences are not required to contain the same medical events, nor the same sequence length. 
Figure~\ref{fig:roc_ages} presents the AUROC for binary LOS prediction stratified by age groups for those with more than 100 patient samples. Interestingly, the model performance is stable for different age groups, with an AUROC of 0.78 for the age groups 60-70 and 70-80 years.
Furthermore, stratification based on the sex of patients yielded similar results, pointing to the robustness of M-BERT for LOS classification.

%% file: sections/conclusion.tex
\section{Conclusion}\label{sec:conclusion}

We have presented a novel approach for predicting LOS by modeling patient information as event sequences. Our approach adapts the transformer machine-learning approach for sequence prediction, which is able to handle the unique features of medical event sequences, namely grouped events and a variety of data types. Our empirical evaluation on a large cohort of emergency care patients from a Danish hospital demonstrates that our model can achieve high accuracy on various LOS problems, 
while outperforming traditional non-sequence machine learning approaches.
Future work could include pre-training of the transformer-based model on a medical task to further improve its performance. Overall, the proposed approach has the potential to improve resource allocation and support decision making in healthcare organizations by providing accurate predictions of LOS. All experimental code is available online\footnoteref{footnote:online_appendix}.




%




